\documentclass[mlmain,twocolumn]{jmlr}

\usepackage{longtable}
\usepackage{subcaption}
\usepackage{booktabs}
\usepackage[load-configurations=version-1]{siunitx} 
\usepackage{float}
\usepackage{multirow}
\usepackage{adjustbox}

\theorembodyfont{\upshape}
\theoremheaderfont{\scshape}
\theorempostheader{:}
\theoremsep{\newline}

\jmlrvolume{}
\firstpageno{1}

\jmlryear{2023}

\newcommand{\hanyu}[1]{\textcolor{black}{#1}} 
\usepackage{soul}

\title[ECG-SL: Electrocardiogram(ECG) Segment Learning]{ECG-SL: Electrocardiogram(ECG) Segment Learning, a deep learning method for ECG signal}

 \author{\Name{Han Yu} \Email{hy29@rice.edu}\\
 \addr Department of Electrical and Computer Engineering, Rice University
 \AND
 \Name{Huiyuan Yang} \Email{hyang@mst.edu}\\
 \addr Department of Computer Science, Missouri University of Science \& Technology
 \AND
 \Name{Akane Sano} \Email{akane.sano@rice.edu}\\
 \addr Department of Electrical and Computer Engineering, Rice University
 }

\begin{document}

\maketitle

\begin{abstract}
Electrocardiogram (ECG) is an essential signal in monitoring human heart activities. Researchers have achieved promising results in leveraging ECGs in clinical applications with deep learning models. However, the mainstream deep learning approaches usually neglect the periodic and formative attribute of the ECG heartbeat waveform. In this work, we propose a novel ECG-\textbf{S}egment based \textbf{L}earning (\textbf{ECG-SL}) framework to explicitly model the periodic nature of ECG signals. More specifically, ECG signals are first split into heartbeat segments, and then structural features are extracted from each of the segments. Based on the structural features, a temporal model is designed to learn the temporal information for various clinical tasks. Further, due to the fact that massive ECG signals are available but the labeled data are very limited, we also explore self-supervised learning strategy to pre-train the models, resulting significant improvement for downstream tasks.
The proposed method outperforms the baseline model and shows competitive performances compared with task-specific methods in three clinical applications: cardiac condition diagnosis, sleep apnea detection, and arrhythmia classification. Further, we find that the ECG-SL tends to focus more on each heartbeat's peak and ST range than ResNet by visualizing the saliency maps. 
\end{abstract}

\vspace{-3mm}
\section{Introduction}
Electrocardiogram (ECG), one of the fastest and most common measures of human heart activities, has a recognized high value in clinical applications. For instance, the waveform in ECG has been proven to be associated with diseases such as cardiovascular diseases \citep{jain2014fragmented}, sleep apnea \citep{faust2016review}, and Parkinson's disease \citep{haapaniemi2001ambulatory}.

As one of the most proven and widely-used computational methods in inspecting ECG, researchers delicately engineered and extracted features from ECG data and trained machine learning models with the crafted features for specific tasks \citep{varon2015novel, venkatesan2018ecg, sharma2019automated}. These works achieved promising performance.
With the great success of deep models in various domains, researchers started embracing the end-to-end deep learning models on ECG signals. The widely used deep models include: i) temporal backbone models, such as long short-term memory (LSTM) \citep{gao2019effective, wang2019global, hou2019lstm} and the Transformer \citep{natarajan2020wide}, ii) structural backbone models, such as convolutional neural networks (\textit{i.e., CNN \citep{baloglu2019classification}, ResNet \citep{jing2021ecg}}), and iii) the combination of both, for example, the combination of ResNet and LSTM \citep{zhou2020ecg}, have achieved promising results in various tasks.

\begin{figure*}[t]
  \centering 
  \includegraphics[width=0.85\textwidth]{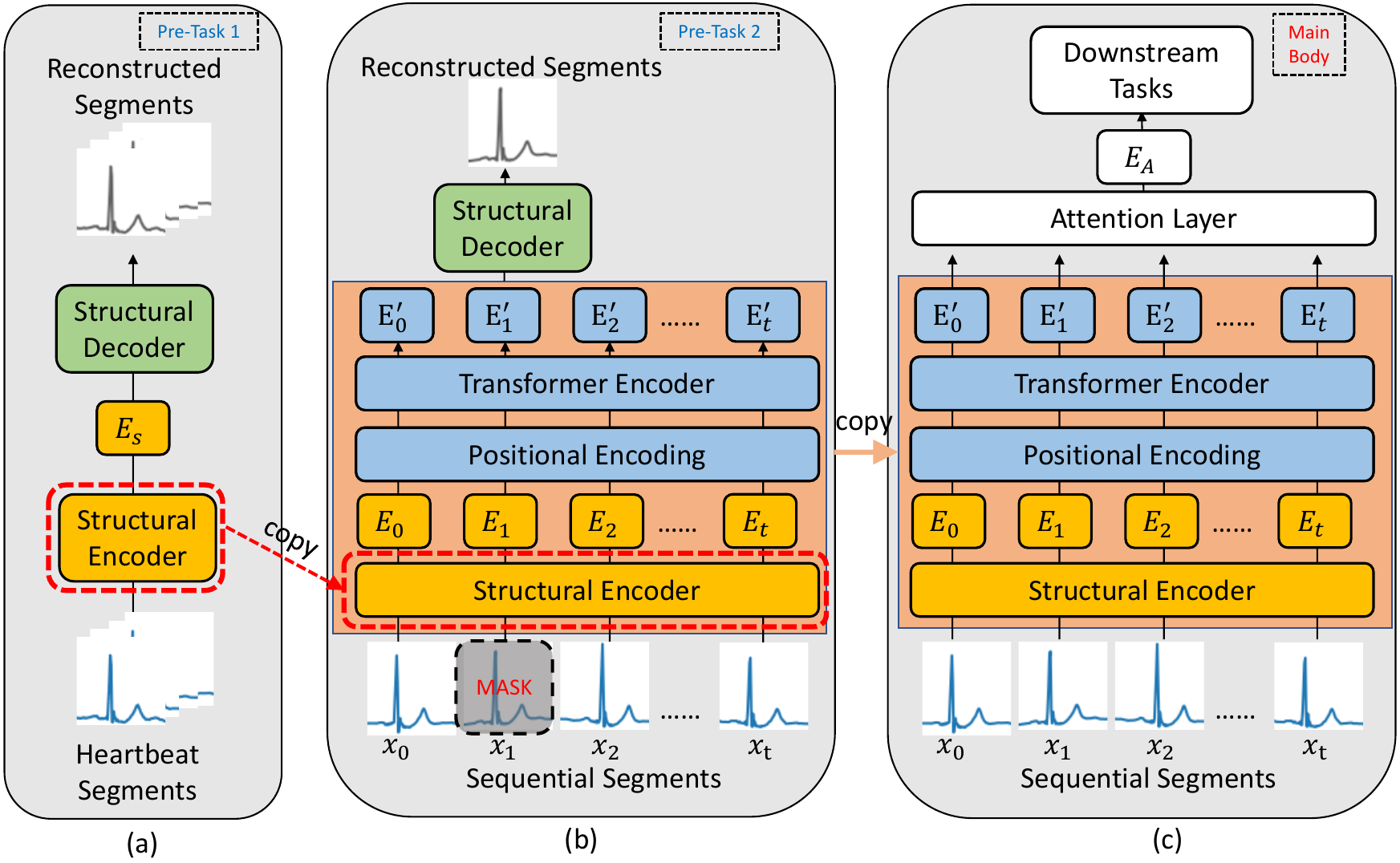}
  \caption{The proposed ECG-SL framework. (c) shows them main model architecture. The input ECG signals are segmented into separate heartbeats to extract structural embedding. The temporal embedding are learned as the model representation for output tasks. A two-stage pre-training task shown as (a) and (b) is operated to help the robustness of the structural and temporal representations}
  \label{fig:ECG_model_framework} 
\end{figure*}

ECG signals are periodic, usually consisting of multiple heartbeat cycles. Vital information is encoded in those formative heartbeat waveforms. Conventional deep learning-based methods (\textit{RNNs, CNNs, etc.}) usually feed the whole ECG signal into a model without considering the nature of heartbeat cycles. The deep models are expected to learn the cardiac cyclicity automatically, but that depends on the availability of massive training data, and it is not guaranteed that deep models learn the cardiac cyclicity, as deep models only implicitly learn the cardiac cyclicity (\textit{if they were able to}). However, the labeled data are usually very limited for clinical tasks, making it challenging for deep models to automatically learn the cardiac cyclicity, and the implicit modeling of the cardiac cyclicity lack of guarantee, therefore sub-optimal for clinical tasks. To boost the performance of deep learning-based models for ECG signal-based clinical tasks, we consider the periodic nature of ECG signals, proposing a segmentation-based deep learning method for ECG data over various clinical tasks. \hanyu{Although the ECG segment-based deep learning method has been proposed in the literature \citep{li2021bat}, the temporal attribute of ECG segments was neglected in the previous study.} Further, we proposed a two-step self-supervised pre-training strategy to learn robust representations by leveraging massive ECG signals. The whole proposed framework is shown in Figure \ref{fig:ECG_model_framework}.


Our contributions can be summarized as: 
\begin{itemize}
    \item We propose a novel segment-based deep learning method for ECG signal to model the cardiac cyclicity implicitly as the structural embedding. On top of the structural embedding, We also model the temporal representation from ECG segments, which was neglected in the previous study. The periodic nature of ECG signals contains vital information but usually neglected by previous works. In this work, we propose a novel ECG-segment based learning framework to explicitly model the periodic nature of ECG signals, which is demonstrated effective in reducing the learning difficulty and  alleviating the over-fitting issue.
    \item 
    To leverage the massive unlabeled ECG signals, we explore and propose a two-stage self-supervised pre-training strategy to learn robust representations for various clinical tasks.
    \item We demonstrate the decision-making mechanism of the proposed ECG-SL method by visualizing the gradient-based saliency map for the input layer. 
\end{itemize}

\vspace{-3mm}
\section{Related Work}
\label{sec:related_work}
ECG is a vital and typical physiological data that monitors the electrical signal of human heart activities and has been widely used in various clinical applications. 
With the rapid development of deep machine learning techniques and the availability of more public ECG datasets, the community is embracing deep models for ECG-based applications.

\vspace{-3mm}
\subsection{Deep Models for ECG}
Due to the temporal attribute of ECG, researchers started using recurrent neural network (RNN) and 1D CNN structures to learn representations from ECG. 
RNN is well-developed for the time-series applications, 
e.g., detecting arrhythmia of heartbeats \citep{gao2019effective, wang2019global, hou2019lstm}. For example, \cite{wang2019global} proposed a global and updatable classification scheme named Global Recurrent Neural Network (GRNN), which improved the generalization of learning from heterogeneous ECG signals. They detected ventricular ectopic beats (VEBs) and supraventricular ectopic beats (SVEBs) with an accuracy of over 99\% on three databases. 

Researchers also applied one-dimension CNN to extract the local information from ECG and aggregated the representation to make predictions for clinical tasks \citep{li2017classification, xiong2017robust, wang2019new, wasserlauf2019smartwatch, urtnasan2018automated, zhu2020classification, hannun2019cardiologist, fang2022sleep}. For instance, \cite{hannun2019cardiologist} applied a 16-block ResNet and achieved the cardiologist-level performance in arrhythmia detection. 
Further, there are also several works in literature that leveraged both the structural models (CNN, ResNet, etc.) and temporal models (RNN, Transformer \citep{vaswani2017attention}, etc.) in learning ECG representations \citep{miao2020continuous, natarajan2020wide, che2021constrained}. \cite{miao2020continuous} employed an 18-layer ResNet structural to extract the local representation from the ECG signal and then aggregated the local information along with an LSTM network to estimate the blood pressure. 

The heartbeat segment-based deep learning approaches has also been proven by other researchers. For example, \cite{li2021bat} applied a SWIN-Transformer \citep{liu2021swin} structure in extracting the embedding of randomly selected five heartbeat segments, then aggregated the learned embedding with the pooling method in diseases diagnosis applications with competitive performances. \hanyu{Nevertheless, they did not leverage the temporal attributes of the ECG segments and aggregated representations of randomly picked heartbeats by summing up embedding.} Differed from their approaches, our proposed method leveraged all the heartbeat segments and aggregated the extracted heartbeat representations in a time-series manner.

\vspace{-3mm}
\subsection{Self-supervised Learning for ECG}
Self-supervised learning (SSL) is a subset of unsupervised learning methods.  SSL refers to methods are proposed to learn general features from large-scale unlabeled data without using any human-annotated labels. A common procedure is to define various pretext tasks for deep models to solve, and the features are learned through this process by learning objective functions defined by the pretext tasks.
ECG-centered self-supervised learning has been proposed by researchers. For example, \cite{sarkar2020self} proposed a data transformation-based self-supervised pre-training tasks to help their model be robust to various signal noise. The proposed method outperformed the baseline methods in mental status classification tasks.  \cite{kiyasseh2021clocs} applied contrastive learning framework for ECG collected from different patients, which aimed to boost the robustness of backbone models among different subjects. Their method outperformed the baselines in classifying cardiac arrhythmia. \hanyu{However, due to the distinct-different model architecture, the previously proposed methods are not compatible to be applied in this work.}
\hanyu{Differing from the aforementioned methods, the proposed two-stage self-supervised learning strategy covers the reconstruction of the masked segments that forced the model to enhance the robustness of the temporal representations.}

\vspace{-3mm}
\section{Methods} \label{sec:methods}
In this section, we introduce the details of the proposed ECG-SL method. As shown in Fig.\ref{fig:ECG_model_framework}, the input ECG signals are first split into $N$ heartbeat segments as $X = [x_{1}, x_{2}, \dots, x_{i}, \dots, x_{N} ]^{T}, x_{i} \in \mathbb{R} ^ {S}, i =\{1, 2, \dots , N \}$, where $S$ is the length of each segment.  The structural embedding of ECG segments is extracted by  1D CNN-based auto-encoder (Fig.\ref{fig:ECG_model_framework}(a)). Based on that, the temporal information is learned by the transformer encoder layers (Fig.\ref{fig:ECG_model_framework}(b)). The pre-trained model is then fine-tuned for various downstream tasks (Fig.\ref{fig:ECG_model_framework}(c)).

\vspace{-3mm}
\subsection{Model Architecture}
\label{sec:model_arch}
The  architecture of the proposed ECG-SL method is shown in Figure \ref{fig:ECG_model_framework}(c). A six-layer convolutional kernel served as the structural encoder to extract embedding $E$ from segmented ECG heartbeats. The extracted structural embedding was combined with positional encoding  as input for the Transformer structure \citep{vaswani2017attention}, which is used to capture the temporal information from the heartbeat sequence. 
Finally, we utilized an attention layer to aggregate all the extracted features and output predictions for various downstream tasks.

\vspace{-3mm}
\subsection{Pre-training ECG-SL}
\label{sec:structural_emb}
The pre-training of the ECG-SL model was two fold. We first pre-trained the structural embedding by leveraging an auto-encoder. Then, inspired by \cite{devlin2018bert}, we developed a bidirectional masked segments reconstruction using the Transformer encoder structure. These two pre-training steps are shown Figure \ref{fig:ECG_model_framework}(a) and (b), respectively. 

\vspace{-3mm}
\subsubsection{Structural Encoder}
\label{sec:stru_enc}
As mentioned in section \ref{sec:model_arch}, the structural encoder was formed with 6 1D CNN layer. We designed an auto-encoder that employed the structural encoder to extract latent space representations $E_s$ from segmented heartbeats $X$. The decoder consisted of 6 1D transposed convolution operators to up-sample the latent representation to the size of the originally inputted heartbeats as $\hat{X}$. The auto-encoder structure was optimized by the mean squared loss between $X$ and $\hat{X}$. The objective function of auto-encoder task can be formulated as:
$$L_{AE}(X) = \|X - \hat{X}\|_2^2$$

\vspace{-3mm}
\subsubsection{Transformer Encoder}
On top of the pre-trained structural encoder discussed in section \ref{sec:stru_enc}, the transformer encoder structure was used to learn temporal representations across segmented heartbeats. Inspired by BERT \citep{devlin2018bert}, we randomly masked out 10\% of the segments $x_m$ among $X$ with all zero tensors. After passing the positional encoding layer and the transformer encoder layers, we reconstructed the masked segments as $\hat{x_i}$. Then, the MSE loss between $x_i$ and $\hat{x_i}$ was used to optimize the model parameters of the transformer structure. Such a design of pre-training task forced the model to learn content from segments across the time axis, thus learning temporal information from the sequence. The objective function can be formulated as:
$$L_{T}(X) =\frac{1}{N} \sum_{i = 1, \dots, N} \|x_i - \hat{x_i}\|_2^2$$
which focused only the masked segments $x_m$ in $X$.

\vspace{-3mm}
\subsubsection{Pre-training Data}
We leveraged ECG from multiple public datasets, including different leads from MIMIC-III waveform \citep{johnson2016mimic}, Apnea-ECG \citep{penzel2000apnea}, PTB-XL \citep{wagner2020ptb}, and PhysioNet Challenge 2017 \citep{clifford2017af}. For the auto-encoder segment reconstruction task,  ~22 million segments were sampled in different leads from MIMIC-III waveform, and ~3 millions segments were sampled from the other three datasets. Besides,  480,000 segments with varied length were sampled from all datasets to perform the masking reconstruction task. 


\vspace{-3mm}
\subsection{Fine-tuning ECG-SL}
Based on the structural and temporal embedding that extracted by the pre-trained architecture, we applied a attention layer, which operated weighted mean reduction across the  representations outputted by transformer encoder for aggregating information. Then, a non-linear dense layer was used to generate the predictions for down-stream tasks.

\vspace{-3mm}
\subsection{Model Interpretation: Saliency Map}
Interpretability of a deep learning model enables human users to understand the reasoning behind predictions and decisions made by the model, which is critical in clinical applications. 
In this work, we also aimed to provide the visualization of the proposed ECG-SL method to help understand its inference mechanism and the difference compared to other baseline methods. We used the gradient-based saliency map \citep{simonyan2013deep} on the top of our models to highlight the critical region of the input ECGs. By tracking the model inference backward through the model, we calculated the gradients in the first input layer, which were used in visualizing the region importance of input ECGs. 

\vspace{-4mm}
\section{Experimental Settings}
\label{sec:exp_res}
This section presents the experimental settings. The data pre-processing procedures including filtering and ECG segmentation can be found in Appendix \ref{apd:preprocessing}. To compare our method with traditional methods such as LSTM, CNN, ResNet, as well as a previous ECG segment-based model BAT \citep{li2021bat}, we implemented baseline models as in Appendix \ref{apd:baseline_models}.

The proposed framework was evaluated  on 3 downstream tasks, including cardiac condition diagnosis, sleep apnea detection, and arrhythmia classification.
Considering the label imbalance in the target datasets, we applied macro f1-score as the evaluation metrics in all tasks as the model performance. The macro f1-score was obtained by computing the average value of the f1-score for each class. In addition, to compare the results with state-of-the-art (SOTAs), we measured the sensitivity (Sen) and specificity (Spe) for sleep apnea detection. Similarly, we monitored the macro precision (prec) and macro recall scores for arrhythmia classification.

\vspace{-3mm}
\subsection{Cardiac Condition Diagnosis}
ECG is an essential tool to diagnose patients' cardiac conditions. We applied our method on the Physikalisch-Technische Bundesanstalt (PTB)- Extra Large (XL) dataset \citep{wagner2020ptb} to detect 5 conditions across 18885 patients. The detailed description of the PTB-XL dataset and train/test sample split can be found in Appendix \ref{data:ptbxl}. Since the current version of the ECG-SL was pre-trained in a single-lead manner, we leveraged only Lead-V ECG (100 Hz) in PTB-XL, although all 12 leads are available for both 100 Hz and 500 Hz sampling frequencies. In the selection of fine-tuning hyperparameters, we used a batch size of 64 and fine-tuned the model for 20 epochs over all the training samples with Adam optimizer and a learning rate of (1e-4).

\vspace{-3mm}
\subsection{Sleep Apnea Detection}
Apnea is one kind of sleep disorder prevalent in male adults world-widely. To model sleep apnea detection with ECG, we leveraged the Apnea-ECG dataset \citep{penzel2000apnea}, which contains single lead ECG data and minute-by-minute binary apnea labels. The detailed description of the Apnea-ECG dataset and train/test sample split is included in Appendix \ref{data:apnea-ecg}. In the selection of fine-tuning hyperparameters, we fine-tuned for 20 epochs with a batch size of 64 over all the training samples using Adam optimizer and a learning rate of (5e-4).

\vspace{-3mm}
\subsection{Arrhythmia Classification}
We conducted experiments on the PhysioNet (CinC) Challenge 2017 \citep{clifford2017af} for classifying ECG signals of normal sinus rhythm, atrial fibrillation, and other arrhythmias. Since the original test set is not accessible to the public, we operate a five-fold cross-validation split and monitor the average scores for this task. The detailed dataset descriptions and train/test split information can be referred to Appendix \ref{data:physionet2017}. In the selection of fine-tuning hyperparameters, we used a batch size of 64 and fine-tune the model for 20 epochs over all the training samples with Adam optimizer and a learning rate of (1e-4).

\vspace{-4mm}
\section{Results}
This section shows the evaluation results for three applications. For the performance of each class in different applications, the confusion matrix of the pre-trained ECG-SL method for each task can be found in Appendix \ref{apd:class_performance}.

\vspace{-3mm}
\subsection{Cardiac Condition Diagnosis}
Table \ref{tab:disease_diag} shows the evaluation performances of predicting 5 cardiac conditions on the test set with metrics of accuracy and macro F1 score. Without pre-training, the proposed model achieved a sightly higher macro f1 score compared to the baselines; whereas with the pre-training initialization, the ECG-SL method outperformed all baseline methods by a notable margin, obtaining at least 4.41\% and 0.097 in accuracy rate and macro f1 score, respectively. Also, we listed the performance of applying the ResNet baseline model with the whole 12-lead ECG. The proposed method shows close results even with only one lead input signal. However, due to the differences in experiment settings, such as the number of leads, the comparison with SOTAs was not conducted for this task. 

\begin{table*}[t]
  \centering 
  \caption{Evaluation results in cardiac disease diagnosis. The bold values represent the highest performance among the results conducted with single-lead ECG.}
  \vspace{-2mm}
\begin{tabular}{c|cc} \hline
Methods                     & Accuracy     & F1$_{macro}$                   \\ \hline
LSTM                      & 50.14   & 0.2800         \\
CNN                       & 69.73   & 0.4849         \\
ResNet                    & 70.93   & 0.4919      \\
ResNet + LSTM             & 70.11   & 0.4761        \\
BAT \citep{li2021bat}     & 70.86   & 0.4983      \\ \hline
ECG-SL (random initialization)         & 71.00   & 0.5016               \\
ECG-SL (pre-trained)         & \textbf{75.10}   & \textbf{0.5742}  \\ \hline\hline
ResNet(12 leads)                    & 77.41   & 0.6011      \\ \hline
\end{tabular}
\vspace{-2mm}
\label{tab:disease_diag}
\end{table*}

\vspace{-3mm}
\subsection{Sleep Apnea Detection}
The experimental results on the Apnea-ECG dataset are shown in Table \ref{tab:res_sleep_apnea}. By training the ECG-SL model from scratch, we observed lower sensitivity and a same-level marco f1 score compared to the baseline CNN models. However, when initializing the backbone with pre-trained parameters, the proposed method outperformed all the baseline models by substantial margins on sensitivity, specificity, and macro f1 scores. The difference in downstream tasks from self-supervised pre-training also reflected in the training effectiveness. For example, we achieved the results in the table by 20 training epochs for the pre-trained parameters vs. 40 epochs for training from scratch. The performance of the proposed method was also comparable to the SOTA methods in the literature, as it showed higher resulting specificity than 3 SOTA works listed in the table.

\begin{table*}[t]
  \centering 
  \caption{Evaluation results in detecting sleep apnea. The bold values represent the best performance among the baseline models and the proposed methods.}
  \vspace{-0.2cm}
\begin{tabular}{c|ccc}
\hline
Methods                                                 & Sen   & Spe   & F1$_{macro}$    \\ \hline \hline
Multi-Scale ResNet \citep{fang2022sleep} & 0.841 & 0.871 & -               \\
1D CNN \citep{chang2020sleep}            & 0.811 & 0.920 & -               \\
CNN + Decision Fusion \citep{singh2019novel} & 0.900 & 0.838 & -               \\
Auto-Encoder + HMM \citep{li2018method} & 0.889 & 0.821 & -               \\\hline
LSTM                                                     & 0.753 & 0.695 & 0.6133          \\
CNN                                                      & 0.856 & 0.784 & 0.8024          \\
ResNet                                                   & 0.843 & 0.774 & 0.7911          \\
ResNet + LSTM                                            & 0.833 & 0.788 & 0.7998          \\
BAT \citep{li2021bat}     & 0.807   & 0.763  & 0.7733   \\ \hline
ECG-SL (random initialization)   & 0.826 & 0.795 & 0.8067  \\ 
ECG-SL (pre-trained)                               & \textbf{0.871} & \textbf{0.843} & \textbf{0.8498}  \\ \hline
\end{tabular}
\label{tab:res_sleep_apnea}
\end{table*}

\vspace{-3mm}
\subsection{Arrhythmia Classification}
Tabel \ref{tab:res_arrhythmia} shows the evaluation performance of the baseline methods and the SOTAs, including the methodology that optimized explicitly for this application \citep{lee2021learning, krasteva2021application}. The pre-trained ECG-SL outperformed the baseline models, as well as the random initialized ECG-SL with 50 training epochs, by wide margins. Also, our results show competitive performance compared to the specialized systems in the literature.

\begin{table*}[t]
  \centering 
  \caption{Evaluation results in classifying arrhythmia. The bold values represent the best performance among the baseline models and the proposed methods.}
  \vspace{-0.2cm}
\begin{tabular}{c|ccc}
\hline
Methods                                                 & Prec$_{macro}$   & Recall$_{macro}$   & F1$_{macro}$    \\ \hline \hline
ResNet \citep{andreotti2017comparing} & - & - & 0.756               \\
2D CNN + Spectrogram \citep{zihlmann2017convolutional}            & - & - & 0.792               \\
DenseNet + Spectrogram \citep{krasteva2021application} & 0.809 & 0.797 & 0.802               \\
BIT-CNN \citep{lee2021learning} & 0.829 & 0.807 & 0.8175               \\\hline
LSTM                                                     & 0.700 & 0.684 & 0.6965          \\
CNN                                                      & 0.751 & 0.744 & 0.7473          \\
ResNet                                                   & 0.756 & 0.748 & 0.7528          \\
ResNet + LSTM                                            & 0.750 & 0.743 & 0.7467          \\
BAT \citep{li2021bat}     & 0.743   & 0.735  & 0.7381   \\\hline
ECG-SL (random initialization)         & 0.767 & 0.750 & 0.7558  \\
ECG-SL (pre-trained)                               & \textbf{0.810} & \textbf{0.796} & \textbf{0.8019}  \\ \hline
\end{tabular}
\vspace{-0.2cm}
\label{tab:res_arrhythmia}
\end{table*}

\vspace{-4mm}
\section{Discussion} \label{sec:discussion}
The designed experiments demonstrated the effectiveness of the proposed ECG-SL method. To develop a further understanding of the proposed method, this section covers the ablation studies, saliency map visualization, and the limitations of this study.

\vspace{-3mm}
\subsection{Ablation Studies}
\subsubsection{ECG Segmentation}
ECG segmentation and the processing of sliced segments are the basis of this work. We tried different strategies for segmenting ECG. For example, as the build-in function in Neurokit package \citep{Makowski2021neurokit}, we tried stretching the captured ECG segments on the pivot of heartbeat peaks to a fixed length instead of performing edge padding as mentioned in section \ref{sec:seg_padding}. This operation caused low performances, which were 65.86\% and 0.4437 in accuracy and macro f1 score, for the preliminary supervised experiments of cardiac condition diagnosis. We suspect that the stretching of ECG segments may distort the rhythm information of ECG heartbeat for specific tasks. Also, we tried padding segments with zeros rather than the edge values. However, the observed reconstruction loss of the training auto-encoder with zero padding was higher than using edge padding (2e-3 vs. 3e-4 in MSEs).

\vspace{-3mm}
\subsubsection{Pre-training Strategy}
Pre-training the structural and temporal embedding is essential to gain robust model representation. We tried to pre-train the mask segments reconstruction task from scratch with the transformer encoder layers; however, it turned out that the reconstructed segments were not close to the original masked ones, with an MSE of 0.06 in average. This fact was our motivation of pre-training the structural embedding with auto-encoder. Regarding structural embedding, we tried pre-training the auto-encoder structure with 20 million segments only in lead-II from the MIMIC-III dataset. Nevertheless, the pre-trained auto-encoder model struggled to recover heartbeat segments in the different leads in other datasets, e.g., Apnea-ECG, as a generalization issue. The MSEs increased from 3e-4 to 8e-3 instantly. Then, by pre-training the structural embedding with the segment in different datasets that cover varying leads, the generalization issue has been resolved.


\vspace{-3mm}
\subsection{Saliency Visualization}

\begin{figure}[!ht]
    \centering
        \includegraphics[width=0.8\columnwidth]{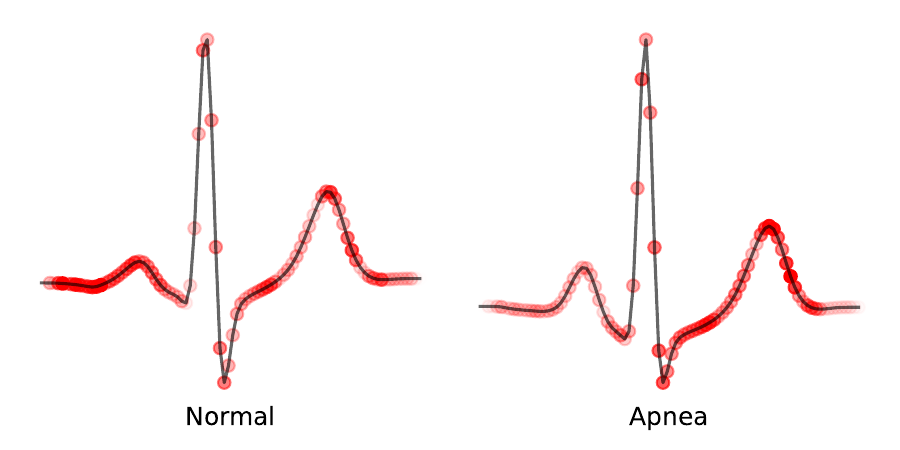}
        \vspace{-2mm}
        \\
       \includegraphics[width=0.8\columnwidth]{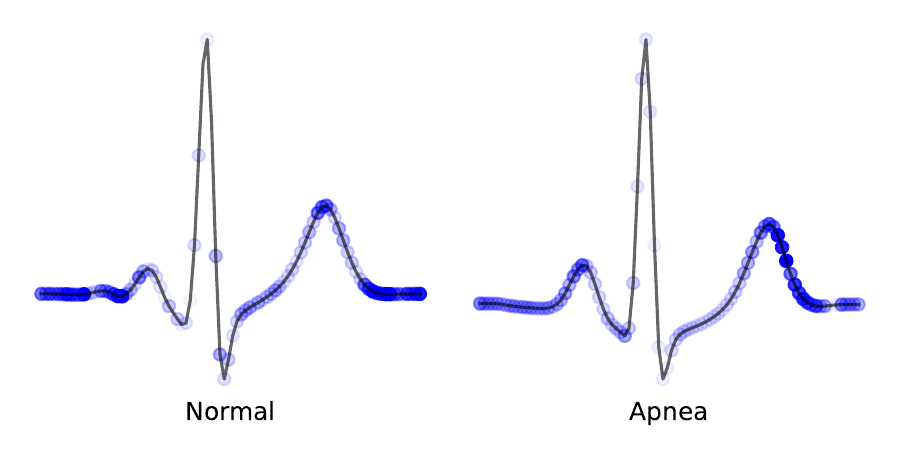}
        \vspace{-2mm}
    \caption{The averaged saliency map on the averaged heartbeat segments of each predicted class for ECG-SL (a) and baseline ResNet model (b) in sleep apnea detection. The intensity of the red and blue dots represent the averaged saliency for (a) and (b), respectively.}
    \label{fig:saliency_map}
    \vspace{-2mm}
\end{figure}

\begin{figure}[!ht]
    \centering
        \includegraphics[width=0.8\columnwidth]{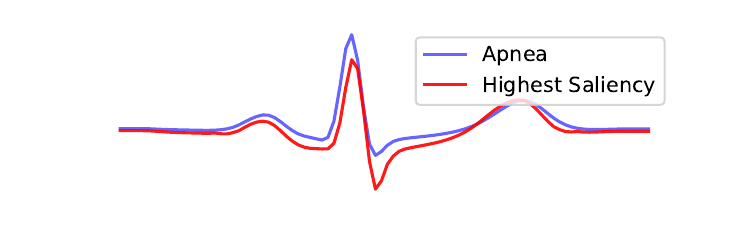}
    \vspace{-4mm}
    \caption{The averaged saliency map on the averaged heartbeat segments of the detected apnea and with the highest saliency.}
    \label{fig:seg_high_saliency}
    \vspace{-5mm}
\end{figure}

To provide the summarized explainable insights upon the input signals, we averaged heartbeat segments and the corresponding saliency maps for each predicted class in the test set. For example, Figure \ref{fig:saliency_map} shows the averaged saliency maps for normal and apnea for the sleep apnea detection task. Although we did segment the input for ResNet, we aligned the saliency map of the corresponding interval by aligning the peaks for comparing the saliency map with ECG-SL. From the figure, the ECG-SL showed higher interests on the P wave when detecting the normal case; whereas higher saliency was spotted on the T wave ans S-T interval while detecting the apnea case. In contrast, the ResNet model focused more on the T-P interval, with more concentration on the offset of T wave while detecting the apnea case. The legend of the aforementioned ECG fiducial points can be found in Appendix \ref{apd:ecg_waveform}.

Further, for each sample that detected as the apnea class, we indexed and averaged the segment with the highest saliency values as shown in Figure \ref{fig:seg_high_saliency}. Compared to the averaged segment for the whole apnea class, the averaged segment on the high saliency showed lower waveform, especially in the QRS region.

\vspace{-3mm}
\subsection{Limitation}
This work also has some limitations. For example, we do not fully capture the complete information of the T-P interval when segmenting the heartbeats. With segmented signals, the waveform of T-P interval is not observable, which might cause information loss from raw ECG signals. 

In addition, there is potential of improving method performance by tuning experimental details such as model hyperparameters, pre-processing procedures, etc. For example, we monitored lower specificity scores in sleep apean detection tasks when comparing our baseline CNN with the 1D CNN work implemented by \citep{chang2020sleep}, with the similar model structure.


\vspace{-3mm}
\section{Conclusion} \label{sec:conclusion}
In this study, we proposed a deep learning method for learning ECG with heartbeat-based structural embeddings and temporal representations, called ECG-SL. We designed a two-stage self-supervised learning strategy to pre-train the model embedding. The method was evaluated on three public datasets for varying clinical tasks, and performance improvements were observed compared to the baseline models. Also, we demonstrated the decision-making mechanism of the proposed method by showing gradient-based saliency maps. In the future, we aim to extend this work for more than a single lead and test our approach in more applications.


\bibliography{pmlr-sample}

\clearpage
\appendix
\section{ECG Waveform} \label{apd:ecg_waveform}
\begin{figure*}[!ht]
  \centering 
  \includegraphics[width=0.85\textwidth]{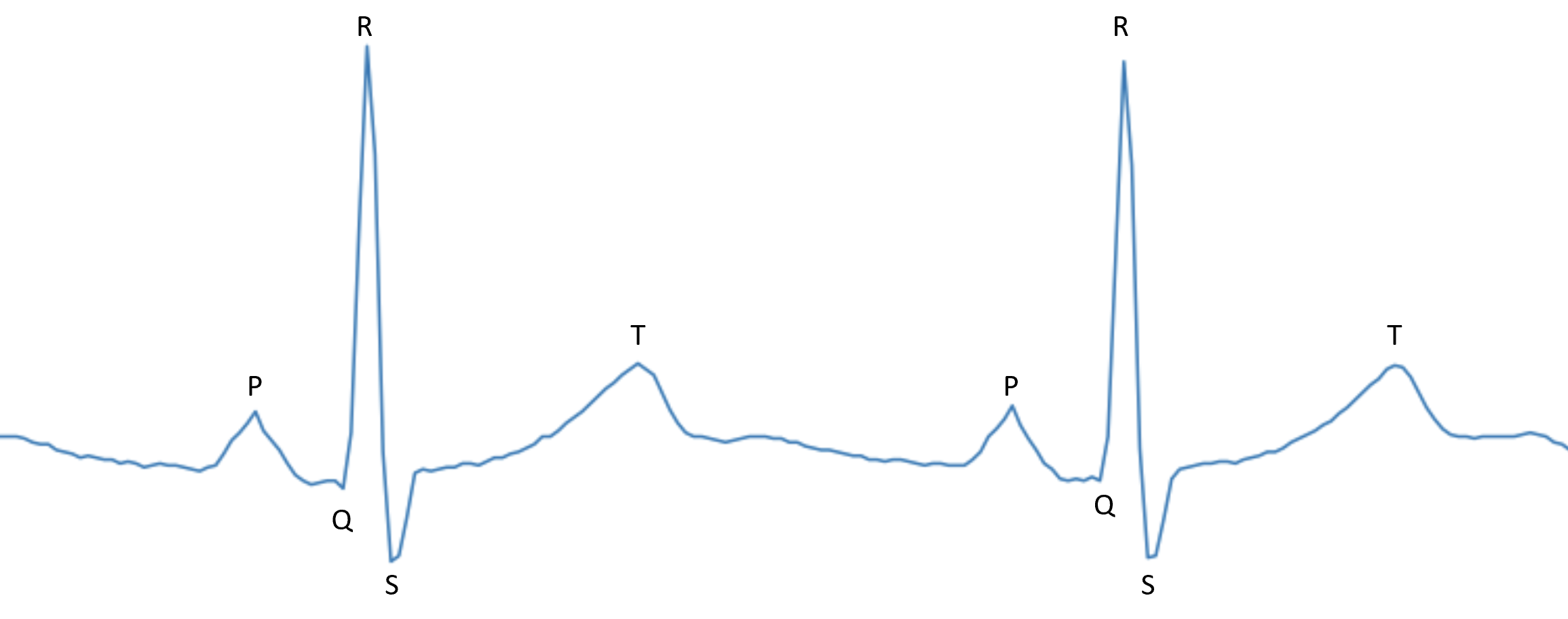}
  \caption{The example waveform and fiducial points of ECG}
  \label{fig:ecg_fiducial} 
\end{figure*}

The example ECG waveform with the annotation of fiducal points can be learned in Figure \ref{fig:ecg_fiducial}.

\section{Experiment Settings}\label{apd:experiment_setting}

\subsection{Datasets}\label{apd:dataset}

\subsubsection{Data Set Meta Information Table}\label{apd:data-meta}
The meta information about the three datasets used in this paper is summarized in Table \ref{dataset_meta}.
\begin{table*}
\centering
\caption{Meta information about datasets used for evaluation}

\begin{tabular}{c|ccc}
\hline 

Task               & \begin{tabular}[c]{@{}c@{}}Cardiac \\ condition \\ diagnosis\end{tabular} & \begin{tabular}[c]{@{}c@{}}Sleep \\ apnea \\ detection\end{tabular} & \begin{tabular}[c]{@{}c@{}}Arrhythmia\\  classification\end{tabular} \\ \hline \hline
Dataset                  & PTB-XL                                                                    & Apnea-ECG                                                           & PhysioNet CinC \\ &&&Challenge 2017                                                                 \\\hline
Training Set Size  & 12976                                                                     & 17233                                                               & 6822                                          \\ \hline
Test Set Size      & 1652                                                                      & 17010                                                               & 1706                                            \\ \hline
\# of Classes      & 5                                                                         & 2                                                                   & 3                                                                      \\ \hline
ECG Duration (s)   & 10                                                                        & 60                                                                  & 9-61                                                          \\ \hline
\# of Participants & 18885                                                                     & 70                                                                  & -                  \\ \hline
\end{tabular}
\label{dataset_meta}
\end{table*}

\subsubsection{Cardiac Condition Diagnosis: PTB-XL} \label{data:ptbxl}
The Physikalisch Technische Bundesanstalt (PTB) comprises a public database, the PTB diagnostic extra large (PTB-XL) \citep{wagner2020ptb}. The dataset was accessed from Physionet \citep{goldberger2000physiobank}. The PTB-XL dataset contains 21837 clinical 12-lead ECG sequences from 18885 patients of 10-second length with two available sampling rates of 100 and 500 Hz. Two cardiologists annotated the sequences with 5-class labels. These five classes are normal (NORM), conduction disturbance (CD), myocardial infarction (MI), hypertrophy (HYP), and ST/T change (STTC). Also, we followed the strategy of splitting training and test sets by \cite{wagner2020ptb}. We excluded the samples which belong to more than one diagnostic category in our experiments. Under this scenario, there are 12976, 1642, and 1652 samples in training, validation, and test set, respectively. And the portions of labeled classes in training set for NORM, CD, MI, HYP, and STTC are 56.10\%, 10.32\%, 15.27\%, 3.18\%, and 15.13\%, respectively.

\subsubsection{Sleep Apnea Detection: Apnea-ECG} \label{data:apnea-ecg}
The Apnea-ECG dataset studies the relationship between human sleep apnea symptoms and the heart activities (monitored by ECG) \citep{penzel2000apnea}. This database can be accessed through Physionet \citep{goldberger2000physiobank}. This dataset contains 70 records with a sampling rate of 100 Hz, from where 35 records were divided into training, and the other 35 were divided into the test set. The duration of the records varies from slightly less than 7 hours to nearly 10 hours. The labels were the annotation of each minute of each recording indicating the presence or absence of sleep apnea. Thus, we split the ECG recording into each minute, which was a total of 6000 data points for each separation. We extracted 17233 samples for the training set and 17010 samples for the test set. And the ratio of non-apnea and apnea samples in the training set was 61.49\% to 38.51\%.

\subsubsection{PhysioNet CinC Challenge 2017} \label{data:physionet2017}
This dataset was originally released for the PhysioNet CinC Challenge 2017 to model the arrhythmia classification task by using ECG signals. The dataset contains 8528 single-lead fingertip ECG recordings in 300 Hz with the duration from 9 to 61 seconds (30 seconds on average). The ECG recordings were annotated by experts into the following classes: normal sinus rhythm (normal, 5076 samples), atrial fibrillation (AF, 758 samples), and abnormal rhythms that do not belong to AF (other, 2415 samples), and noisy signals. In this study, we downsample the sequences into 100 Hz, and we mainly focused on the classification among normal, AF, and other classes. Since the original evaluation set is not released to public, we performed a stratified five-fold cross-validation following \citep{lee2021learning}.


\subsection{Pre-training Data} 
\label{apd:pretrain_data}
In this study, the data samples for pre-training tasks are all sampled from 4 datasets: MIMIC-III waveform (M), Apean-ECG (A), PTB-XL (P), and PhysioNet CinC Challenge 2017 (C). We sample around 25 million heartbeat segments from all datasets to pre-training the structural embedding with an auto-encoder. They consist of (1) 22 million segments from randomly sampled 300,000 one-minute ECG segments evenly in leads I, II, III, and V; (2) 1 million segments from PTB-XL training set in all 12 leads; (3) 1.5 million segments from the training set of Apnea-ECG data and (4) 0.5 million segments from the PhysioNet CinC challenge 2017 dataset.

In the masked segment reconstruction task, we sampled (1) 300,000 one-minute ECG sequences evenly from the MIMIC-III waveform dataset; (2) 155,712 10-second samples from PTB-XL in 12 different leads; (3) 17,233 one-minute samples from Apnea-ECG and (4) 8528 varying length samples from PhysioNet CinC Challenge 2017 dataset. In total, we used around 480 thousand ECG sequences in the masked segment reconstruction task.

\subsection{Data Processing} \label{apd:preprocessing}
\subsubsection{Filtering and Segmentation}
Raw ECG signals usually contain noisy components due to numerous factors. Therefore, we designed a noise filter by first i) applying a Butterworth filter with a low-cut frequency of 0.5Hz to filter out the low-frequency noise and then ii) using a moving average kernel with a width of 50Hz to filter out the 50Hz powerline (natural noise). The filtering operations were implemented with Neurokit \citep{Makowski2021neurokit}. Also, considering the data compatibility across datasets, we resampled all the signals with a sampling rate of 100 Hz. 

Further, we performed segmentation based on the peak finder in Neurokit \citep{Makowski2021neurokit} with the de-noised ECG signal. The peak of each heartbeat was located and fixed at the center position of each segmented frame. The segmenting algorithm then captured P and T waves by chunking the 0.35 and 0.45 of the RR-interval before and after the located peak, respectively. Each heartbeat segment was normalized in the range of 0-1.

\subsubsection{Padding}
We utilized two padding operations on the segmented heartbeats to make our data fit the deep learning frameworks. The first \textit{segment padding} operation was to make all segmented heartbeats the same length. Secondly, we used \textit{sequential padding} to ensure each batch of input data has the same time step, which is required for the model training process. 

\textbf{Segment Padding.} \label{sec:seg_padding}
Due to the varied length of the RR-intervals, the heartbeat segments were usually not of the same length. We padded the segments with edge values of each waveform so that the length segment reached a constant size of $S$. In this study, $S$ was set as 100.

\textbf{Sequential Padding.} 
The number of heartbeat segments varied with different ECG signals. For instance, 60-second ECG sequences could contain 50 to over 150 heartbeat segments depending on subjects and confounder factors. 
We padded the shorter sequences with zero segments so that each sequence was in the same time steps within the input data batch.

\subsection{Baseline Models}
\label{apd:baseline_models}
Different models have been used as the baseline, including LSTM, CNN, ResNet, BAT, and the combination of ResNet and LSTM. This subsection introduces the details of our implementation. 
\subsubsection{LSTM}
Three layers of bidirectional LSTM are stacked as the basic temporal representation extractor. The filter size for all 3 LSTM layers is set as 32, and the output from the last cell of the 3-rd LSTM layer is used as the learned representation from ECGs. Two dense layers with neurons of 512 and 256 were used to map the representations from LSTM to the numbers of classes (varied in different tasks).

\subsubsection{CNN}
We followed a similar CNN structure as used in \citep{urtnasan2018automated}, which used six convolutional layers with an filtered ECGs. During our implementation, the filter sizes for six convolutional layers were [32, 32, 64, 64, 128, 256], and the kernel size for each layer was set as 5. The max-pooling layer was applied after every convolutional layer to reduce the dimension of representations. Instead of a fully connected layer, an adaptive average pooling layer was used to make the designed model adaptive to different ECG input lengths for various tasks. 
Finally, the output layer mapped the representations to the predictions for a specific application.

\subsubsection{Beat-aligned Transformer (BAT)}
The BAT model \citep{li2021bat} is also a segment-based method that applied the Swin Transformer \citep{liu2021swin}, which is a variant of the visual Transformer, to extract structural information from randomly selected heartbeat segments. The extracted representations are aggregated by an average pooling layer. We implemented this method based on the original released code base. According to the applications, we randomly selected 50\% of the heartbeat segments to represent each inputting ECG sequence. On top of the heartbeat segment, we leveraged the Swin Transformation module to extract the representation from heartbeats and used an average pooling layer to aggregate all the information for the prediction.

\subsubsection{ResNet}
ResNet is a widely used backbone for various computer vision tasks, which has been even adapted to model time series data and achieved promising performance.
During our experiment, we implemented the 1D version of the ResNet18 \citep{he2016deep}. There were eight residual blocks in total, and each block contained two convolutional layers and one batch normalization layer. 
The filter number of convolutional layers in the first residual block was set as 64, and then the filter number doubled every two residual blocks.
Thus, we expected representations in 512 channels as the output of the last residual block. Same as the baseline CNN, the kernel size used in ResNet is also 5. And we also employed an adaptive average pooling layer after residual blocks to aggregate the learned representations. Finally, the input and output dimensions of the last dense layer were 512 and numbers of classes, respectively.

\subsubsection{ResNet + LSTM}
Based on our implementations of LSTM and ResNet. Our implementation in the combination of these two models was straightforward. Same as the ResNet model, we first used the ResNet structure to extract the representations from the whole ECG inputs. However, we abandoned using the adaptive averaging pooling layer and the fully connected layer after residual blocks. The LSTM was responsible in connected the extracted representations. The input dimension of the LSTM cell was 512 as the embedding from residual blocks was in 512 channels. Still, we used the output from the last cell of the 3-rd LSTM layer as the learned representation. A dense layer with 512 neurons was used to output the prediction results. 

\section{Confusion Matrix for ECG-SL} \label{apd:class_performance}

\begin{figure}[H]
  \centering 
  \includegraphics[width=0.45\textwidth]{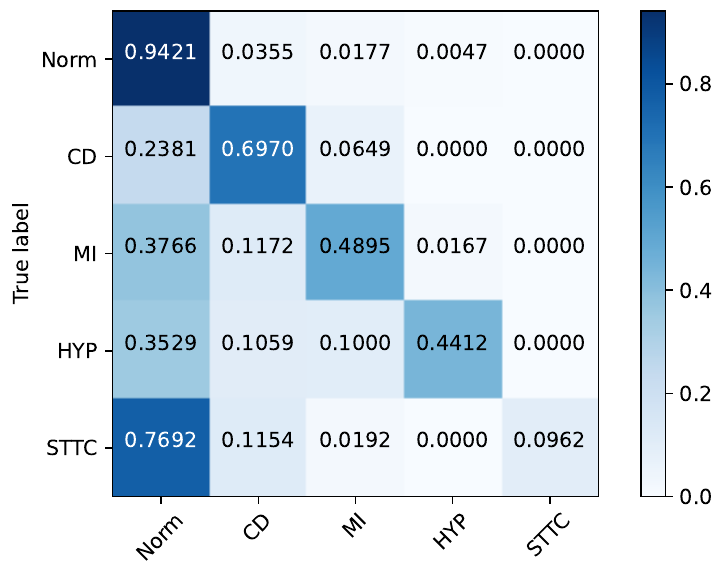}
  \caption{The confusion matrix resulted from ECG-SL in cardiac disease diagnosis}
  \label{fig:cm_ptbxl} 
\end{figure} 

\begin{figure}[H]
  \centering 
  \includegraphics[width=0.45\textwidth]{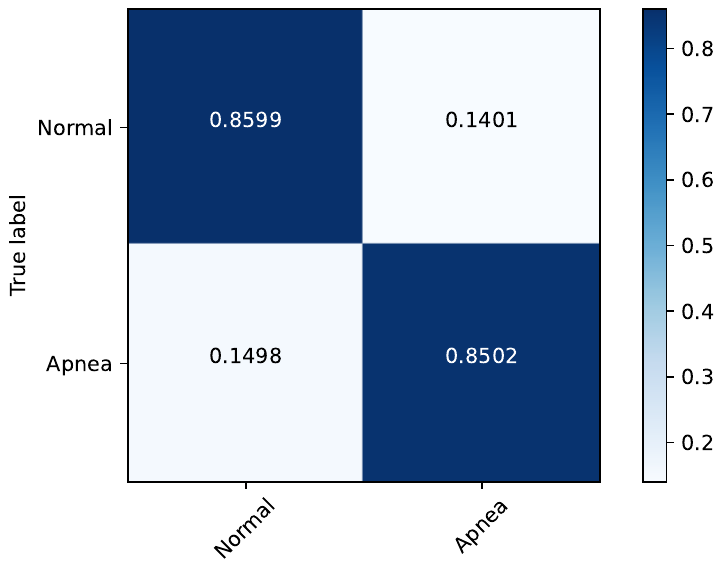}
  \caption{The confusion matrix resulted from ECG-SL in sleep apnea detection}
  \label{fig:cm_apnea} 
\end{figure} 

\begin{figure}[H]
  \centering 
  \includegraphics[width=0.45\textwidth]{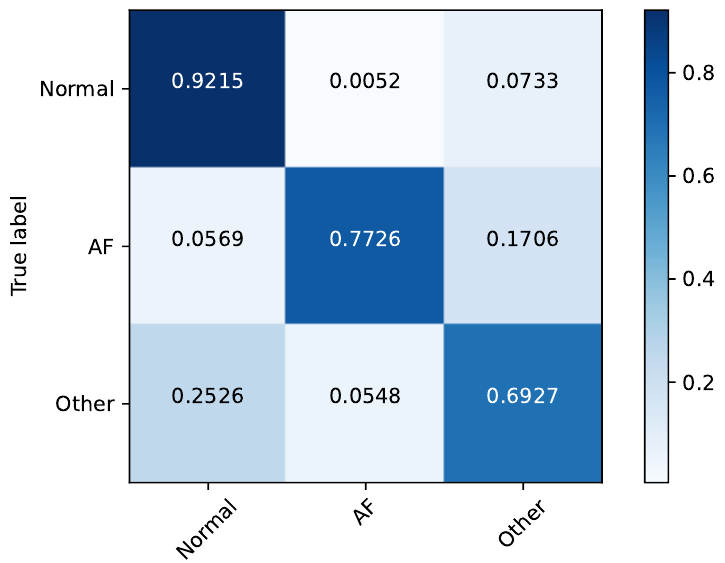}
  \caption{The confusion matrix resulted from ECG-SL in Arrhythmia detection}
  \label{fig:cm_arrhythmia} 
\end{figure}

\end{document}